\documentclass[conference]{IEEEtran}
\IEEEoverridecommandlockouts

\usepackage{cite}
\usepackage{multirow}
\usepackage{amsmath,amssymb,amsfonts}
\usepackage{algorithmic}
\usepackage{graphicx}
\usepackage{textcomp}
\usepackage{xcolor}
\def\BibTeX{{\rm B\kern-.05em{\sc i\kern-.025em b}\kern-.08em
    T\kern-.1667em\lower.7ex\hbox{E}\kern-.125emX}}

\usepackage{fancyhdr}
\begin{document}

\title{Assessing the Impact of Personality on Affective States from Video Game Communication

\thanks{This work is funded by James S McDonnell Foundation (Grant Title: A Methodology for Studying the Dynamics of Resilience of College Students).}
}

\author{\IEEEauthorblockN{Atieh Kashani}
\IEEEauthorblockA{\textit{Department of Computational Media} \\
\textit{University of California, Santa Cruz}\\
Santa Cruz, USA\\
atkashan@ucsc.edu}
\and
\IEEEauthorblockN{Johannes Pfau}
\IEEEauthorblockA{\textit{Department of Computational Media} \\
\textit{University of California, Santa Cruz}\\
Santa Cruz, USA\\
jopfau@ucsc.edu}
\and
\IEEEauthorblockN{Magy Seif El-Nasr}
\IEEEauthorblockA{\textit{Department of Computational Media} \\
\textit{University of California, Santa Cruz}\\
Santa Cruz, USA\\
mseifeln@ucsc.edu}
}

\maketitle
\thispagestyle{fancy}

\maketitle
\begin{abstract}
Individual differences in personality determine our preferences, traits and values, which should similarly hold for the way we express ourselves. With current advancements and transformations of technology and society, text-based communication has become ordinary and often even surpasses natural voice conversations - with distinct challenges and opportunities. In this exploratory work, we investigate the impact of personality on the tendency how players of a team-based collaborative alternate reality game express themselves affectively. We collected chat logs from eleven players over two weeks, labeled them according to their affective state, and assessed the connection between them and the five-factor personality domains and facets. After applying multi-linear regression, we found a series of reasonable correlations between (combinations of) personality variables and expressed affect - as increased confusion could be predicted by lower self-competence (C1), personal annoyance by vulnerability to stress (N6) and expressing anger occured more often in players that are prone to anxiety (N1), less humble and modest (A5), think less carefully before they act (C6) and have higher neuroticism (N). Expanding the data set, sample size and input modalities in subsequent work, we aim to confirm these findings and reveal even more interesting connections that could inform affective computing and games user research equally.

\end{abstract}


\section{Introduction}
Communication is a complex subject that can be influenced by numerous factors including individual differences and their emotional or affective states. During a communication act, individuals express affects in different ways by their choice of words, facial expression, vocal features, gesture and body language. Both verbal and nonverbal cues play an important role in the way that affect is expressed and interpreted through communication\cite{859332,scherer1984vocal}. With the rise of digital media, communications are increasingly performed using text-based computer-mediated communication. The lack of nonverbal cues in mediated communication has led to the assumption that text-based communication has a reduced capacity for exchanging affective states \cite{kiesler1986hidden}. However, text-based communication can convey various ranges of emotions and affects by adapting the forms that are distinct from those found in non verbal communication \cite{hancock2007expressing,walther2005let}. During text-based communication, communicators encode the emotions and affects that they would normally communicate through nonverbal cues into the other forms such as emoticons, deformed spellings, punctuation, acronyms and special abbreviations\cite{walther2001impacts,walther1992interpersonal,thurlow2003generation}. In addition, synchronous real-time text communication can capture some of the synchronicity that is associated with voice or face-to-face communication \cite{Scott2012AdaptingGT}. Thus, the emergence of real-time and online communication platforms has created new avenues for studying the verbal behaviour phenomenon and its psychological correlates.

Individual differences refer to the variations that exist between humans with regard to personality, cognition and behaviours. Personality has been defined as "a stable, organized collection of psychological traits and processes in the human being that influences his or her interactions with and modifications to the psychological, social and physical environment surrounding them" \cite{larsen2005personality}. The different personality traits can manifest in various ways including how individuals experience and express affects or emotions in verbal communication. The Five Factor Model (FFM) \cite{digman1990personality} is the most accepted and widely used personality theory that provides a systematic assessment of emotional, interpersonal, experiential, attitudinal, and motivational styles. While the five overarching domains of the FFM are too broad to capture the complex human personality in detail, underlying individual facets form a more precise description of personality to differentiate between individuals and their behaviours, including the expression of affects \cite{costa1995domains}.   

Video games have the potential to place individuals in the continuous mode of interaction that evoke emotional and affective responses. Players are drawn to play games not only for enjoyment and achieving rewards but also for engaging in experiences that may even elicit negative emotions like sadness, anxiety, and frustration\cite{granic2014benefits}. Game features such as mechanics, interactive gameplay, storyline and immersive graphics make them a unique platform in affective computing research\cite{yannakakis2014emotion} for studying psychological constructs and social phenomena. In particular, Alternative Reality Games (ARG) can construct a close connection to reality, as they embed players in a fictional narrative that unfolds through interaction with real-world applications, such as mobile phones, text messages and social networks \cite{10.1145/1314215.1314222}. In ARGs, the interactions and in-game events often mimic real life situations that can engage participants for an extended period of time. Utilizing ARGs allows researchers to incorporate engaging and ecologically valid methods to study various aspects of human behavior by capturing multi-dimensional data on humans' interactions and communications.    

Altogether, this creates a unique opportunity to study the impact of personality on verbal expression of affects set in ARG-mediated communication, which we approach in this work. Investigating the connection between personality and expression of affect can lead to several potential benefits such as more inclusive design, adaptive personalization and tailored interventions through understanding individual differences. Thus, we formulate our research endeavor into the following research question:

\begin{itemize}
    \item Can we identify connections between individual personality differences and the tendencies to express oneself in distinct affective categories from in-game chat conversation?
\end{itemize}

By exploring and presenting initial relations between personality and affect expression through game communication, we contribute to games user research and affective computing.

\section{Related Work}
Previous studies show that the expression of emotions or affects in conversation varies as a function of individual differences and personality traits \cite{holtgraves2011text,fast2008personality,Komulainen2014TheEO,mehl2003sounds, berry1997linguistic}. Holtgraves investigated the correlations between the five-factor model of personality (extroversion, neuroticism, agreeableness, conscientiousness, and openness to experience) and how it impacted the use of language in text messaging \cite{holtgraves2011text}. He reported that increased neuroticism was associated with the more frequent occurrence of negative emotion words, higher scores on extroversion were associated with the occurrence of fewer negative words, and agreeableness was negatively correlated with the use of negative emotion words. Another study also found that agreeableness is positively correlated with the occurrence of positive emotion words and negatively with negative emotion words \cite{berry1997linguistic}. Komulainen et al. reported that conscientiousness positively associate with positive affect and negatively associate with negative affect \cite{Komulainen2014TheEO}. Consistent with previous findings, recent studies show that individuals high on self-reported extroversion tend to use more positive emotion words \cite{chen2020meta} and individuals high in conscientiousness demonstrate their prudence by refraining from expressing negative emotions \cite{tackman2020personality}. 

For different application purposes, Volkmar et al. tailored in-game achievements to individual differences and measured an increase of player experience if matching properly \cite{volkmar2019player}. Teng et al. used player journey map segmentation to investigate differences in gameplay based on -- or influencing -- higher-level metrics, which are not limited to personality variables \cite{teng2022player,pfau2023player}. Habibi et al. measured differences in physiological responses between different personalities, especially higher impact of stress on more extroverted persons \cite{habibi2023under}. In subsequent work, they also predicted individual personality differences from low-level in-game behavior and (pre-defined) communication choices \cite{habibi2023modeling}, which were yet far from unconstrained, natural speech.

The mentioned studies considered only five factors of personality traits and none of them examined how facets manifested in natural text-based communication. 
In addition, these studies reported the impact of personality traits on the broad emotional or affective states (positive and negative). Therefore, the specific and discrete expression of emotions and affects were not examined. 
 
The current study attempts to address those limitations by utilizing a serious ARG to examine how the occurrence of expression of distinct affects in verbal communication varies as a function of individuals' personality traits and facets. 

\section{Study}
To situate our investigation into a suitable Alternate Reality Game, we draw on the game called \textit{LUX} \cite{habibi2022data}, which was developed and the data were collected by a group of researchers and developers with the aim of measuring resilience and coping strategies in first-year undergraduate students. \textit{LUX} is a multiplayer team-based cooperative game designed to foster communication within solving complex puzzles and challenges. It is set in a fictional narrative and takes place and interacts with the real world, while presenting challenges and stressors to assess emotional and affective responses. The game is composed of multiple episodes, and each episode consists of a series of puzzles in which players need to communicate with a bot and other team members through Discord in order to solve them. We collected the players' chat data, identified affective states throughout the messages and linked them to their self-reported FFM personality variables (cf. Section \ref{measures}). 

\subsection{Measures}
For measuring the participants' personality, we have utilized the Revised NEO Personality Inventory (NEO PI-R) \cite{costa1995domains} as the standard self-report questionnaire measure of the Five Factor Model (FFM), which provides a systematic assessment of emotional, interpersonal, experiential, attitudinal, and motivational styles. The NEO PI-R is a concise measure of the five major domains of personality (Neuroticism, Extraversion, Openness, Agreeableness, and Conscientiousness). Within each of those broader domains, six specialized traits (facets) together represent a given domain score (which add up to 30 facets in total). 

For qualitative classification of the chat data, we have employed Plutchik's Wheel of Emotion \cite{plutchik1980general} and the taxonomy of affects as discussed by \cite{Scott2012AdaptingGT} to develop a set of labels.  In total, we ended up with ten labels as outlined in Table \ref{tab:BaselineTable}. This label set was then utilized to label the players' conversation. 
\label{measures}. We also considered a ``no affect" label to exclude messages that do not express any affective state. For this investigation, one researcher served as an annotator for all players' conversations. They considered the impact of the situational context and surrounding messages in the conversation to apply the code that captures the affect expressed in the message. They also took into account the impact of verbal cues such as emojis, slang and abbreviations that influenced the affective meaning of the entire sentence. In some cases, these did express actual affects, or were used to avoid misleading others (e.g. after being sarcastic). A total number of 3748 lines of utterances has been labeled line by line, with up to one affective label each.

\begin{table*}[h]
\centering
\begin{tabular}{|c|p{8cm}|p{5cm}|}
\hline
Label & Description  &  Chat data example\\
\hline
Agreement & Being consistent with or accepting other player's opinions or plans towards finding clues and solving puzzles & \textit{``alright sounds good''} \\
\hline
Confusion & Lack of certainty when encountering a situation associated with the game or puzzle & \textit{``I'm not sure if the scans are allowed to be connected to each other ): because so far none of them have..''}\\
 \hline
Disagreement & Lack of consistency in players' opinions or plans in the game or during the process of solving puzzles & \textit{``really don't think it's Octavian''} \\
 \hline
Excitement & Feeling eagerness towards a progress or an intense response as a result of an accomplishment in the game & \textit{``LETS GOOO''} \\
 \hline
Frustration & Feeling of dissatisfaction or hopelessness as a result of failure, being stuck, encountering problems or inabilities to find clues or solve puzzles & \textit{``unfortunately, i just get a bunch of garble''} \\
 \hline
Amusement & Finding an incident funny during game play or in the process of solving puzzles & \textit{``hey, work smarter, not harder lol''} \\
 \hline
Supportive & Being collaborative and providing help to assist the teammates to find clues or solve puzzles & \textit{``I'll scan that when I get home''} \\
 \hline
Annoyance & Feeling unpleasant or irritated as a result of failure or inability to find clues or solve puzzles & \textit{``ugh is this ispy?''} \\
 \hline
Interest & Feeling of wanting or giving attention to the situation associated with the game or puzzle & \textit{``i want to make sure i get through it, so id like to schedule''} \\
 \hline
Anger & Intense emotional response as a result of failure, being stuck, or encountering with a problem in the game in the process of solving puzzles & \textit{``Damn we broke the bot''}
\\
\hline
\end{tabular} \\~\\
\caption{ Set of affect labels from Plutchik's wheel of emotions that appeared within the conversation of the game, together with their definition and an example from the data. }
\label{tab:BaselineTable}
\end{table*}

\subsection{Procedure}
We recruited participants through an on-boarding event on campus, where they agreed to informed consent and data collection. We then asked participants to form a team with three members to start the game, resulting in five teams in total: four teams with three members each and one team with four members. A total of 16 players played the game through two weeks of playtesting, and submitted a post-study questionnaire containing the discussed metrics afterwards. Yet, one team with four members did not finish the game and another single player failed to submit the NEO personality questionnaire, which we excluded from further investigation.
From the remaining eleven players, six identified as male and five identified as female, distributed into four teams.

\section{Results}
We applied a multi-linear regression model using Python Scikit-learn library and analyzed the data to predict the occurrence of each affect based on a personality domain/facet or a  combination of up to four personality domains/facets. To evaluate the results of the prediction, we calculated the Mean Square Error (MSE) values for each personality facet/domain combination. Table \ref{tab:LinearRegressionTable} shows the combination of personality domains/facets that can predict the conversational affect occurrence with highest accuracy (lowest MSE) after five-fold cross-validation. In addition, we calculated the coefficient associated with each particular personalty domain/facet to assess the direction and effect size onto the expressed affect. To benchmark these outcomes against a control condition, we  considered two baselines that follow assumptions that personality would have no impact on the prediction of affect. In the first, the possibility of the occurrence of each affective state in the conversation is equal for all the affects. Considering the ten different affective states in our sample, the probability of occurrence of each affect states in conversation is thus 10\%. The second baseline acknowledges that different affective states are differently likely to appear in the data and is thus constructed based on the mean of the total occurrence of each affect in our sample (cf. columns BL1 and BL2, respectively). Since the number of the “Supportive” affect label in the players’ conversation is higher than the other affect labels, the naïve BL1 and BL2 would show especially high MSE in contrast.

The results showed that the combination of four personality domains/facets predict the conversational affect occurrence with highest accuracy (lowest MSE) on the testing set. We included the top three combinations together with their coefficient towards the affective state. For example, when predicting the occurrence of ``Anger", the combination of ``Anxiety", ``Modesty", ``Deliberation" and ``Neuroticism" had a comparatively low MSE of 0.84, as compared to the two baselines ($MSE_{BL1}=60.6$ and $MSE_{BL2}=9.8$). When taking the coefficients into account as well, personalities with higher ``Modesty" and ``Deliberation" indicated less expressions of ``Anger", while higher ``Anxiety" and ``Neuroticisim" correlated rather positively with ``Anger".

\begin{table*}[h!]
\centering
\begin{tabular}{|c|r r r r |c|c|c|}
\hline
Affect & \multicolumn{4}{c|}{Personality Domains/Facets (and Coefficients)} & \begin{tabular}[x]{@{}c@{}}MSE\\(LR)\end{tabular} & \begin{tabular}[x]{@{}c@{}}MSE\\(BL1)\end{tabular} & \begin{tabular}[x]{@{}c@{}}MSE\\(BL2)\end{tabular} \\
\hline
 & O4: Actions (1.34) & A1: Trust (-1.57) & C1: Competence (0.83) & N: Neuroticism (-0.19) & 1.63 &
 \multirow{3}{*}{21.4} & \multirow{3}{*}{13.1} \\
Agreement & C1: Competence (-0.84)& C3: Dutifulness (-1.16)& 
C4: Achievement (-0.99) & O: Openness (0.64)
 & 2.54 & & \\
& O4: Actions (1.18)& A1: Trust (-1.13)& C3: Dutifulness (-0.27)& N: Neuroticism (-0.12) & 3.04 & & \\
\hline
 & E5: Excitement (-0.77)& C1: Competence (-1.96)& N: Neuroticism (-0.47)& O: Openness (0.74) & 4.16
 & \multirow{3}{*}{54.6} & \multirow{3}{*}{50.4} \\
Confusion & O6: Values (4.47)& A3: Altruism (8.37)& A6: Tender-Mind. (-2.5)& E: Extraversion (-0.66) & 4.60 & &\\
& N6: Vulnerability (-1.98)& O3: Feelings (1.87)& A4: Complicated (0.88)& C1: Competence (-3.03) & 6.9 & & \\
\hline
 & E5: Excitement (0.26)& O2: Aesthetics (0.14)& O5: Ideas (-0.40)& O6: Values (0.74) & 0.11 &
  \multirow{3}{*}{80.4} & \multirow{3}{*}{2.9} \\
Disagreement & E2: Gregariousness (-0.30)& E3: Assertiveness (0.53)& O1: Fantasy (0.73)& O4: Actions (-0.19) & 0.16 & & \\
& N2: Anger Hostility (0.10)& O5: Ideas (-0.32)& O6: Values (0.63)& E: Extraversion (0.09) & 0.17 & & \\
\hline
 & N2: Anger Hostility (-1.0)& A3: Altruism (0.92)& C2: Order (-0.57)& A: Agreeableness (-0.24) & 0.78 &  \multirow{3}{*}{44.1} & \multirow{3}{*}{9.9} \\
Excitement & N2: Anger Hostility (-0.9)& E2: Gregariousness (0.78)& C2: Order (-0.78)& A: Agreeableness (-0.14)  & 0.81 & & \\
& N2: Anger Hostility (-0.62)& E2: Gregariousness (0.82)& C1: Competence (-0.35)& C2: Order (-0.82) & 0.89 & & \\
\hline
  & E6: Pos. Emotions (1.02)& O2: Aesthetics (0.37)& A2: Straightfwd. (-0.33)&  A3: Altruism (0.7) & 0.10 &  \multirow{3}{*}{45} & \multirow{3}{*}{8.6} \\
Frustration & E6: Pos. Emotions (0.83)& O2: Aesthetics (0.38)& A3: Altruism (0.78)& A6: Tender-Mind. (-0.31) & 0.47 & & \\
& E6: Pos. Emotions (0.91)& O2: Aesthetics (0.43)& A3: Altruism (0.87)& A: Agreeableness (-0.12) & 0.44 & &\\
\hline
 & E2: Gregariousness (2.21)& C1: Competence (-1.06)& C2: Order (-2.61)&  A: Agreeableness (0.41) & 3.08 &  \multirow{3}{*}{55.6} & \multirow{3}{*}{58.4} \\
Amusement & N1: Anxiety (0.97)& N2: Anger Hostility (-1.91)& C2: Order (-1.26)& C5: Self-Discipline (-1.55) & 3.48 & & \\
& E3: Assertiveness (1.35)& A1: Trust (-1.36)& C2: Order (-2.17)& A: Agreeableness (0.80) & 4.32 & & \\
\hline
  & C1: Competence (4.6)&  N: Neuroticism (0.72)& O: Openness (2.36)& C: Conscientious (1.2) & 15.63 & \multirow{3}{*}{1551.3} & \multirow{3}{*}{401.3} \\
Supportive & N3: Depression (3.66)& A1: Trust (1.45)& A5: Modesty (7.48)& O: Openness (-1.99) & 20.33 & & \\
& E1: Warmth (-7.51)& O4: Actions (-2.93)& C1: Competence (8.66) & C6: Deliberation (6.33) & 34.72 & &\\
\hline
 & N6: Vulnerability (1.48)& E1: Warmth (-2.96)& A2: Straightfwd. (1.41)& A5: Modesty (-1.29) & 3.04 & \multirow{3}{*}{60.5} & \multirow{3}{*}{25.7}\\
Annoyance & E5: Excitement (1.55)& E6: Pos. Emotions (-1.21)& O3: Feelings (-1.55) & O6: Values (1.16) & 3.80 & &\\
& N5: Impulsiveness (-1.18) & E2: Gregariousness (1.76)& O3: Feelings (-2.07)& C4: Achievement (1.95) & 6.99 & &  \\
\hline
  & E6: Pos. Emotions (-0.35)& O1: Fantasy (0.78)& A5: Modesty (0.78)& C5: Self-Discipline (-0.13) & 0.12 & \multirow{3}{*}{57.5} & \multirow{3}{*}{3.5} \\
Interest & E1: Warmth (0.26)& E6: Pos. Emotions (-0.39)& O1: Fantasy (0.79)& A5: Modesty (0.66) & 0.14 & & \\
& E6: Pos. Emotions (-0.32)& O1: Fantasy (0.76)& A4: Complicated (0.09)& A5: Modesty (0.70)  & 0.17 & &\\
\hline
 & N1: Anxiety (0.35) & A5: Modesty (-1.86)& C6: Deliberation (-1.81)& N: Neuroticism (0.25) & 0.84 & \multirow{3}{*}{60.6} & \multirow{3}{*}{9.8} \\
Anger & O5: Ideas (-0.43)& O6: Values (0.97)& A6: Tender-Mind. (-0.59)& C4: Achievement (0.32) & 0.87 & & \\
& N6: Vulnerability (-0.36)& O6: Values (0.47)& C4: Achievement (0.86)& A: Agreeableness (-0.39) & 0.9 & & \\
\hline
\end{tabular}
~\\
\caption{Most accurate personality domains/facets to predict affective conversation after multi-linear regression (indicating coefficients and mean squared errors), as compared to a baseline that assumes every label would appear equally (BL1) and a baseline that includes the different distributions of affects in conversation, but neither regards the impact of personality on the prediction (BL2).}
\label{tab:LinearRegressionTable}
\end{table*}

\section{Discussion}
When interpreting the results of the prediction (as summarized in Table \ref{tab:LinearRegressionTable}), certain relationships could be identified that are arguably reasonable, while others do not necessary align with the background literature, or display inconclusive results, which we outline in the following. Although we attempted to support our findings with relevant psychological literature, we encountered a lack of sufficient research in some areas. Therefore, we proceeded with interpreting the outcomes. 

``Confusion'' was best predicted by the personality facets of E5, C1, N and O. The strong negative correlation between a person's perceived self-competence (C1) and the probability to be confused by a logical puzzle of the game seems coherent. Also, as this method does not necessarily measure confusion itself but rather the chance of expressing that one is confused (in comparison to other personalities), it is reasonable that this willigness to admit one's own confusion goes along with higher Openness (O) and lower Neuroticism (N) in general.

Individuals who are more vulnerable to stress (N6) also expressed to be ``Annoyed'' more often. Same holds for people with less emotional warmth (E1) and less modesty or humbleness (A5), which definitely stands to reason. This is only underlined by the positive correlation to straight-forward personalities (A2), as they are arguable less likely to withhold their frustration.

``Anger'' is positively correlated with the proneness to anxiety, worriedness and nervousness (N1), which make up reasonable predictors for this.  This is similarly justifiable as with the connection to players that are less humble and modest (A5), think less carefully before they act (C6) and have higher Neuroticism (N) in general.

The likelihood of expressing more ``Supportive'' statements is highly correlated with an individual's perceived self-competence (C1), as players could probably give better support when understanding the current challenge themselves. The connections to high openness towards other people (O) and high conscientousness (C) similarly play well into this, while a positive connection to Neuroticism (N) remains at least debatable.

The tendency to experience anger, frustration and bitterness (N2) consistently correlates negatively with players that expressed their ``Excitement'' more often, which arguably makes sense. Same might hold for personalities that highly value other people's welfare and experience (A3) or tend to be less organized (O2). Yet, one would have hypothesized that the personality facet of seeking excitement (E5) would also have a stronger connection to the expression of ``Excitement'' throughout chat data.

Gregarious people enjoy the company of others (E2), which explains the high correlation with their expressed ``Amusement''. The negative correlation of self-perceived competence (C1) and expressed amusement is however debatable, as there is no simple linear connection between objective ability (or subjective competence) and happiness \cite{veenhoven2012does}. 

The tendency to experience positive emotions (E6) is highly predictive of the players' expressed frustration, which does not compute at first glance, yet the facet does not exclude the experience of negative emotions per se. This correlation could still stand to reason for people that are generally more prone to experience both negative as well as positive emotions, but a connection to facets that particularly target negative emotions would have been more reasonable.

However, at least with regards to background theories, we cannot justify the correlation between ``Disagreement'' and people who are more excitement-seeking (E5), or personalities that have a deep appreciation for art and beauty (O2). In fact, the traits that indicates the openness to accepting new ideas and other opinions (O5) is negatively correlated with ``Disagreement'' in these results, where the opposite would be more intuitive.

Thus, we engage with our introductory research question, arguing that we delivered initial insights that individual personality differences can strongly impact affective expression in game communication, and that most of the derived connections are reasonably justifiable, barring some limitations that are discussed in the following.

\section{Limitations and Future Work}
Altogether, most of the predicted multi-linear correlations stand to reason, with some exceptions that are presumably caused by the highly noisy domain of individual personality. For the sake of brevity, we did not explicate on all, but only the most predictive facet combinations, and leave remaining interpretations open for the reader through Table \ref{tab:LinearRegressionTable}. Certain connections that we hypothesized to be trivially true (such as the tendency of agreeableness (A) and the expressed ``Agreements'', the personal desire for excitement (A5) and the expression of it, or the hostility towards Anger (N2) and its utterance) were not reflected in the prediction. Yet, we only considered the four major factors that could predict the affect expression in the end, while the former still might have had a smaller effect.

In our current labeling process, we only appointed a single annotator to decide affective labels for the particular chat utterances. While this could already show a working trend of the approach that can come up with reasonable results, personal bias might have influenced the classification of the conversations, which is why we are expanding this process in the next iterative step of this work to multiple annotators and a proper assessment of the inter-rater reliability.

An essential part of the noise that led to the  inconclusive parts of the results could be overcome by incorporating a larger data set of participants, which is what we are currently working towards. Especially the highly variational personality data requires a broad range of different personality combinations in order to come to conclusions that are accurate and usable for large-scale applications. Using in-game and conversational behavior from a vast community of players of \textit{Sky: Children of the Light} \cite{sky}, we are striving to scale our approach and investigate if we can extract comparable or even more accurate findings.

The proposed technique is obviously limited in its applicability to domains that incorporate recorded chat communication. This constrains it to multi-player environments, and only those who actively engage humans in natural language conversation. Yet, with the current rise of large language models and increasing use of novel application cases, we are interested to investigate single-player games that embed natural language conversations with non-player characters for narrative, quest or mechanical reasons, and will derive if there are significant differences in  emotional expression when interacting with artificial agents instead of fellow human players.

Eventually, for this proof of concept that reasonably accurate connections between personality and affect expression through chat are derivable,  we only considered a single method for the modeling process. While the outcomes of the multi-linear regression are intuitively understandable, more sophisticated machine learning approaches could have approximated this connection with even more accuracy. Thus, our future work includes the investigation of such models, while we constrain ourselves to techniques with high explainability (such as random forest regression or Bayesian belief networks) to still be able to ground and justify the underlying functions (in contrast to black-box models).

Limitations with respect to the ethical component of using this and related methodologies are further discussed in Section \ref{sec:ethics}.

\section{Conclusion}
Individual personality differences influence how we make decisions, take stances, display emotions and express ourselves. Video games, especially when incorporating or being based on communication, have the opportunity to engage players in conversation, control topics and insert stimuli, record context-sensitive utterances and can even benefit from assessing affective states of their players to tailor content, difficulty or experiences. Thus, this work explored how the personality of players of a multi-player alternate reality game impacted their expression of affective states when solving puzzles and coordinating with their teams. By classifying their communication into affective labels and modeling the role of their Five Factor Model facets towards that, we present initial results that identify a first differentiation between individuals and their expression of affect in text-based communication. We considered ten primary conversational affects from Plutchik's established wheel of emotions and a combination of up to four facets/domains, which often led to reasonable connections between personality and affective expression already. Based on this, we are looking forward to investigate large-scale observations between personality and expression, how to accurately model these in the context of games, and how to make avail of these to tailor player experiences through difficulty, content and matchmaking.

\section{Ethical Statement}
\label{sec:ethics}
The realized study closely followed procedure, framing and informed consent as approved by the institutional review board of the authors' affiliated university.

While this proposed technology aims at opening up understanding individual differences and could tailor game mechanics, environments or matchmaking towards inclusiveness and accessibility, it still bears certain risks and ethical implications that should be addressed. 
First of all, as this approach is working on conversation data which can be highly sensitive and personal, the question of data ownership comes into play. Even if companies provide game environments and services and therefore often have control over incoming and outgoing data, chat data should ideally only be leveraged with the actively confirmed approval from the particular player (i.e. \textit{opt-in}). Ideally, echoing data transparency, players should have full insight and control over the history of their chat logs, so that unwanted entries could be permanently removed from storage and usage for the model. Moreover, even when being able to control their individual input, regular users can hardly estimate the impact of their data and how it could change in-game or higher-level decisions that certain use cases could determine - thus, in the spirit of explainability, users should be able to clearly follow the decisions of the model, its outcomes and implications on their experience with the product.
After all, modeling relationships between chat, personality and affect and (algorithmically) deriving decisions from that should only be deployed for the benefit (e.g. improved experience) of the user, but bears the risk to be exploited to further facilitate dark patterns of (game) design, such as taking advantage of purchasing patterns or reinforcing addictive tendencies.
These risks excel in the case of erroneous decision making of the model, which could steer the individual's experience in the wrong direction or completely spoil it. Thus, if such a model is used for tailoring or adapting any element, it should only do if it can satisfy the prediction following a reasonable confidence.

\section*{Acknowledgment}
LUX was developed and the data were collected by the group of researchers and developers including Reza Habibi, Bjarke Larsen, Sai Siddartha Maram, Shweta Sisodiya, Jonattan Holmes, Zhaoqing Teng, and Jessica Wei at the University of California, Santa Cruz.  

\bibliographystyle{IEEEtran.bst}
\bibliography{bibliography}

\end{document}